\documentclass{article} 
\usepackage{iclr2017_conference,times}
\usepackage{hyperref}
\usepackage{url}
\usepackage{amsfonts}
\usepackage{graphicx}
\usepackage{mathtools}
\usepackage{booktabs}
\usepackage{amsmath}
\usepackage{appendix}
\usepackage{verbatim}

\title{Reinforcement Learning with Unsupervised Auxiliary Tasks}

\author{Max Jaderberg\thanks{Joint first authors. Ordered alphabetically by first name.}~, Volodymyr Mnih\textsuperscript{*}, Wojciech Marian Czarnecki\textsuperscript{*}\\
{\bf Tom Schaul, Joel Z Leibo, David Silver \& Koray Kavukcuoglu}\\
DeepMind\\
London, UK\\
\texttt{\{jaderberg,vmnih,lejlot,schaul,jzl,davidsilver,korayk\}@google.com} \\
}

\newcommand{\argmax}{\operatornamewithlimits{arg\,max}}

\newcommand{\figref}[1]{Figure~\ref{#1}}

\newcommand{\sref}[1]{Section~\ref{#1}}
\def\eg{\emph{e.g.}}
\def\ie{\emph{i.e.}}

\newcommand{\expect}[1]{\mathbb{E} \left[ #1 \right]}
\newcommand{\expectx}[2]{\mathbb{E}_{#1} \left[ #2 \right]}
\newcommand{\dd}[2]{\frac{\partial{#1}}{\partial{#2}}}

\begin{document}

\maketitle

\begin{abstract}
Deep reinforcement learning agents have achieved state-of-the-art results by directly maximising cumulative reward. However, environments contain a much wider variety of possible training signals. In this paper, we introduce an agent that also maximises many other pseudo-reward functions simultaneously by reinforcement learning. All of these tasks share a common representation that, like unsupervised learning, continues to develop in the absence of extrinsic rewards. We also introduce a novel mechanism for focusing this representation upon extrinsic rewards, so that learning can rapidly adapt to the most relevant aspects of the actual task. Our agent significantly outperforms the previous state-of-the-art on Atari, averaging 880\% expert human performance, and a challenging suite of first-person, three-dimensional \emph{Labyrinth} tasks leading to a mean speedup in learning of 10$\times$ and averaging 87\% expert human performance on Labyrinth.
\end{abstract}

Natural and artificial agents live in a stream of sensorimotor data. At each time step $t$, the agent receives observations $o_t$ and executes actions $a_t$. These actions influence the future course of the sensorimotor stream. In this paper we develop agents that learn to predict and control this stream, by solving a host of reinforcement learning problems, each focusing on a distinct feature of the sensorimotor stream. Our hypothesis is that an agent that can flexibly control its future experiences will also be able to achieve any goal with which it is presented, such as maximising its future rewards.

The classic reinforcement learning paradigm focuses on the maximisation of extrinsic reward. However, in many interesting domains, extrinsic rewards are only rarely observed. This raises  questions of what and how to learn in their absence. Even if extrinsic rewards are frequent, the sensorimotor stream contains an abundance of other possible learning targets. Traditionally, unsupervised learning attempts to reconstruct these targets, such as the pixels in the current or subsequent frame. It is typically used to accelerate the acquisition of a useful representation. In contrast, our learning objective is to predict and control features of the sensorimotor stream, by treating them as pseudo-rewards for reinforcement learning. Intuitively, this set of tasks is more closely matched with the agent's long-term goals, potentially leading to more useful representations. 

Consider a baby that learns to maximise the cumulative amount of red that it observes. To correctly predict the optimal value, the baby must understand how to increase ``redness" by various means, including manipulation (bringing a red object closer to the eyes); locomotion (moving in front of a red object); and communication (crying until the parents bring a red object). These behaviours are likely to recur for many other goals that the baby may subsequently encounter. No understanding of these behaviours is required to simply reconstruct the redness of current or subsequent images.

Our architecture uses reinforcement learning to approximate both the optimal policy and optimal value function for many different pseudo-rewards. It also makes other auxiliary predictions that serve to focus the agent on important aspects of the task. These  include the long-term goal of predicting cumulative extrinsic reward as well as short-term predictions of extrinsic reward.
To learn more efficiently, our agents use an experience replay mechanism to provide additional updates to the critics. Just as animals dream about positively or negatively rewarding events more frequently \citep{schacter2012future}, our agents preferentially replay sequences containing rewarding events. 

Importantly, both the auxiliary control and auxiliary prediction tasks share the convolutional neural network and LSTM that the base agent uses to act. By using this jointly learned representation, the base agent learns to optimise extrinsic reward much faster and, in many cases, achieves better policies at the end of training.

This paper brings together the state-of-the-art Asynchronous Advantage Actor-Critic (A3C) framework~\citep{mnih2016asynchronous}, outlined in \sref{sec:background}, with auxiliary control tasks and auxiliary reward tasks, defined in sections \sref{sec:auxcontrol} and \sref{sec:auxreward} respectively. These auxiliary tasks do not require any extra supervision or signals from the environment than the vanilla A3C agent. The result is our UNsupervised REinforcement and Auxiliary Learning (\emph{UNREAL}) agent (\sref{sec:unreal}) 

In \sref{sec:exp} we apply our \emph{UNREAL} agent to a challenging set of 3D-vision based domains known as the \emph{Labyrinth} \citep{mnih2016asynchronous}, learning solely from the raw RGB pixels of a first-person view. Our agent significantly outperforms the baseline agent using vanilla A3C, even when the baseline was augmented with an unsupervised reconstruction loss, in terms of speed of learning, robustness to hyperparameters, and final performance. The result is an agent which on average achieves 87\% of expert human-normalised score, compared to 54\% with A3C, and on average 10$\times$ faster than A3C. Our \emph{UNREAL} agent also significantly outperforms the previous state-of-the-art in the Atari domain.

\begin{figure}[t]
\centering
\vspace{-1.2cm}
\hspace{-5mm} \includegraphics[scale=0.27]{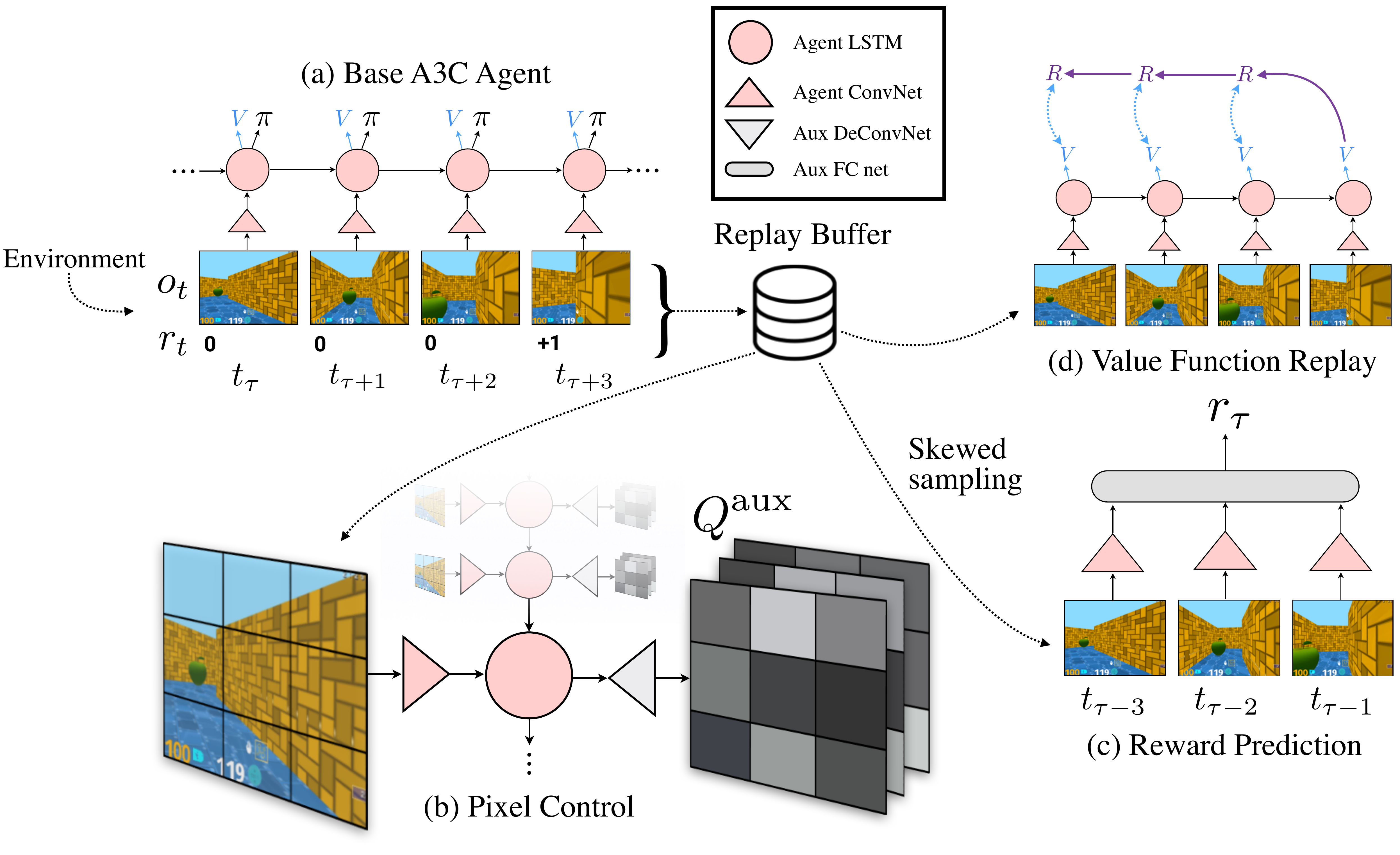}
  \caption{\small Overview of the \emph{UNREAL} agent. (a) The base agent is a CNN-LSTM agent trained on-policy with the A3C loss~\citep{mnih2016asynchronous}. Observations, rewards, and actions are stored in a small replay buffer which encapsulates a short history of agent experience. This experience is used by auxiliary learning tasks. (b) Pixel Control -- auxiliary policies $Q^\mathrm{aux}$ are trained to maximise change in pixel intensity of different regions of the input. The agent CNN and LSTM are used for this task along with an auxiliary deconvolution network. This auxiliary control task requires the agent to learn how to control the environment. (c) Reward Prediction -- given three recent frames, the network must predict the reward that will be obtained in the next unobserved timestep. This task network uses instances of the agent CNN, and is trained on reward biased sequences to remove the perceptual sparsity of rewards. (d) Value Function Replay -- further training of the value function using the agent network is performed to promote faster value iteration. Further visualisation of the agent can be found in \url{https://youtu.be/Uz-zGYrYEjA}}
  \label{fig:overview}
  \vspace{-0.35cm}
\end{figure}

\section{Related Work}

A variety of reinforcement learning architectures have focused on learning temporal abstractions, such as options \citep{sutton:options}, with policies that may maximise pseudo-rewards \citep{konidaris2009skill,silver2012compositional}. The emphasis here has typically been on the development of temporal abstractions that facilitate high-level learning and planning. In contrast, our agents do not make any direct use of the pseudo-reward maximising policies that they learn (although this is an interesting direction for future research). Instead, they are used solely as auxiliary objectives for developing a more effective representation. 

The Horde architecture \citep{sutton2011horde} also applied reinforcement learning to identify value functions for a multitude of distinct pseudo-rewards. However, this architecture was not used for representation learning; instead each value function was trained separately using distinct weights. 

The UVFA architecture \citep{schaul2015universal} is a factored representation of a continuous set of optimal value functions, combining features of the state with an embedding of the pseudo-reward function. Initial work on UVFAs focused primarily on architectural choices and learning rules for these continuous embeddings. A pre-trained UVFA representation was successfully transferred to novel pseudo-rewards in a simple task. 

Similarly, the successor representation \citep{dayan1993improving, barreto2016successor,kulkarni2016deep} factors a continuous set of expected value functions for a fixed policy, by combining an expectation over features of the state with an embedding of the pseudo-reward function. Successor representations have been used to transfer representations from one pseudo-reward to another \citep{barreto2016successor} or to different scales of reward  \citep{kulkarni2016deep}.

Another, related line of work involves learning models of the environment \citep{schmidhuber2010formal,xie2015model,oh2015action}. Although learning environment models as auxiliary tasks could improve RL agents (\eg~\cite{lin1992memory,li2015recurrent}), this has not yet been shown to work in rich visual environments.

More recently, auxiliary predictions tasks have been studied in 3D reinforcement learning environments. \cite{lampe2016doom} showed that predicting internal features of the emulator, such as the presence of an enemy on the screen, is beneficial. \cite{Mirowski16} study auxiliary prediction of depth in the context of navigation.

\section{Background}\label{sec:background}
\vspace{-1em}
We assume the standard reinforcement learning setting where an agent interacts with an environment over a number of discrete time steps.
At time $t$ the agent receives an observation $o_t$ along with a reward $r_t$ and produces an action $a_t$. The agent's state $s_t$ is a function of its experience up until time $t$, $s_t = f(o_1, r_1, a_1, ..., o_t, r_t)$. The $n$-step return $R_{t:t+n}$ at time $t$ is defined as the discounted sum of rewards, $R_{t:t+n} = \sum_{i=1}^{n} \gamma^i r_{t+i}$. The value function is the expected return from state $s$, $V^\pi(s) = \expect{R_{t:\infty} | s_t = s, \pi}$, when actions are selected accorded to a policy $\pi(a|s)$. The action-value function $Q^\pi(s,a) = \expect{R_{t:\infty} | s_t = s, a_t = a, \pi}$ is the expected return following action $a$ from state $s$. 

Value-based reinforcement learning algorithms, such as Q-learning \citep{watkins1989learning}, or its deep learning instantiations DQN \citep{mnih-dqn-2015} and asynchronous Q-learning \citep{mnih2016asynchronous}, approximate the action-value function $Q(s,a;\theta)$ using parameters $\theta$, and then update parameters to minimise the mean-squared error, for example by optimising an $n$-step lookahead loss \citep{peng1996msq}, $\mathcal{L}_Q = \expect{\left( R_{t:t+n} + \gamma^n \ \max_{a'} Q(s',a'; \theta^-) - Q(s,a;\theta) \right)^2}$; where $\theta^-$ are previous parameters and the optimisation is with respect to $\theta$. 

Policy gradient algorithms adjust the policy to maximise the expected reward, $\mathcal{L}_\pi = -\expectx{s\sim\pi}{R_{1:\infty}}$, using the gradient $\dd{\expectx{s\sim\pi}{R_{1:\infty}}}{\theta} = \expect{\dd{}{\theta} \log \pi(a|s) (Q^\pi(s,a) - V^\pi(s))}$ \citep{watkins1989learning,sutton1999policy}; in practice the true value functions $Q^\pi$ and $V^\pi$ are substituted with approximations. The Asynchronous Advantage Actor-Critic (A3C) algorithm \citep{mnih2016asynchronous} constructs an approximation to both the policy $\pi(a|s,\theta)$ and the value function $V(s,\theta)$ using parameters $\theta$. Both policy and value are adjusted towards an $n$-step lookahead value, $R_{t:t+n} + \gamma^n V(s_{t+n+1}, \theta)$, using an entropy regularisation penalty, $\mathcal{L}_\mathrm{A3C} \approx \mathcal{L}_\mathrm{VR} + \mathcal{L}_{\pi} - \expectx{s\sim\pi}{\alpha H(\pi(s, \cdot, \theta)}$, where $\mathcal{L}_\mathrm{VR} = \expectx{s\sim\pi}{\left(R_{t:t+n} + \gamma^n V(s_{t+n+1}, \theta^-) - V(s_t,\theta) \right)^2}$.

In A3C many instances of the agent interact in parallel with many instances of the environment, which both accelerates and stabilises learning. The A3C agent architecture we build on uses an LSTM to jointly approximate both policy $\pi$ and value function $V$, given the entire history of experience as inputs (see \figref{fig:overview} (a)).

\section{Auxiliary Tasks for Reinforcement Learning}

In this section we incorporate \emph{auxiliary tasks} into the reinforcement learning framework in order to promote faster training, more robust learning, and ultimately higher performance for our agents. \sref{sec:auxcontrol} introduces the use of auxiliary control tasks, \sref{sec:auxreward} describes the addition of reward focussed auxiliary tasks, and \sref{sec:unreal} describes the complete \emph{UNREAL} agent combining these auxiliary tasks.

\subsection{Auxiliary Control Tasks}\label{sec:auxcontrol}

The auxiliary control tasks we consider are defined as additional pseudo-reward functions in the environment the agent is interacting with. We formally define an auxiliary control task $c$ by a reward function $r^{(c)}:\mathcal{S}\times\mathcal{A} \rightarrow \mathbb{R}$, where $\mathcal{S}$ is the space of possible states and $\mathcal{A}$ is the space of available actions. The underlying state space $\mathcal{S}$ includes both the history of observations and rewards as well as the state of the agent itself, i.e. the activations of the hidden units of the network. 

Given a set of auxiliary control tasks $\mathcal{C}$, let $\pi^{(c)}$ be the agent's policy for each auxiliary task $c\in \mathcal{C}$ and let $\pi$ be the agent's policy on the base task. The overall objective is to maximise total performance across all these auxiliary tasks,
\begin{equation}
\label{eqn:aux-task-loss}
\argmax_{\theta} \mathbb{E}_{\pi}[R_{1:\infty}] + \lambda_c \sum_{c \in \mathcal{C}} \mathbb{E}_{\pi_c}[R_{1:\infty}^{(c)}],
\end{equation}
where, $R_{t:t+n}^{(c)}=\sum_{k=1}^{n}\gamma^k r_t^{(c)}$ is the discounted return for auxiliary reward $r^{(c)}$, and $\theta$ is the set of parameters of $\pi$ and all $\pi^{(c)}$'s.
By sharing some of the parameters of $\pi$ and all $\pi^{(c)}$ the agent must balance improving its performance with respect to the global reward $r_t$ with improving performance on the auxiliary tasks.

In principle, any reinforcement learning method could be applied to maximise these objectives. However, to efficiently learn to maximise many different pseudo-rewards simultaneously in parallel from a single stream of experience, it is necessary to use off-policy reinforcement learning. We focus on value-based RL methods that approximate the optimal action-values by Q-learning. Specifically, for each control task $c$ we optimise an $n$-step Q-learning loss $\mathcal{L}_Q^{(c)} = \expect{\left( R_{t:t+n} + \gamma^n \ \max_{a'} Q^{(c)}(s',a', \theta^-) - Q^{(c)}(s,a,\theta) \right)^2}$, as described in \cite{mnih2016asynchronous}.

While many types of auxiliary reward functions can be defined from these quantities we focus on two specific types:
\begin{itemize}
\item \textbf{Pixel changes} - Changes in the perceptual stream often correspond to important events in an environment. We train agents that learn a separate policy for maximally changing the pixels in each cell of an $n\times n$ non-overlapping grid placed over the input image. We refer to these auxiliary tasks as \emph{pixel control}. See \sref{sec:exp} for a complete description.
\item \textbf{Network features} - Since the policy or value networks of an agent learn to extract task-relevant high-level features of the environment \citep{mnih-dqn-2015,zahavy2016graying,silver2016mastering} they can be useful quantities for the agent to learn to control. Hence, the activation of any hidden unit of the agent's neural network can itself be an auxiliary reward. We train agents that learn a separate policy for maximally activating each of the units in a specific hidden layer. We refer to these tasks as \emph{feature control}.
\end{itemize}

The \figref{fig:overview} (b) shows an A3C agent architecture augmented with a set of auxiliary pixel control tasks. In this case, the base policy $\pi$ shares both the convolutional visual stream and the LSTM with the auxiliary policies. The output of the auxiliary network head is an $N_\mathrm{act}\times n\times n$ tensor $Q^\mathrm{aux}$ where $Q^\mathrm{aux}(a, i, j)$ represents the network's current estimate of the optimal discounted expected change in cell $(i,j)$ of the input after taking action $a$. We exploit the spatial nature of the auxiliary tasks by using a deconvolutional neural network to produce the auxiliary values $Q^\mathrm{aux}$.

\subsection{Auxiliary Reward Tasks}\label{sec:auxreward}

\begin{figure}[t]
\centering
\vspace{-1.2cm}
\hspace{-5mm} \includegraphics[scale=0.3]{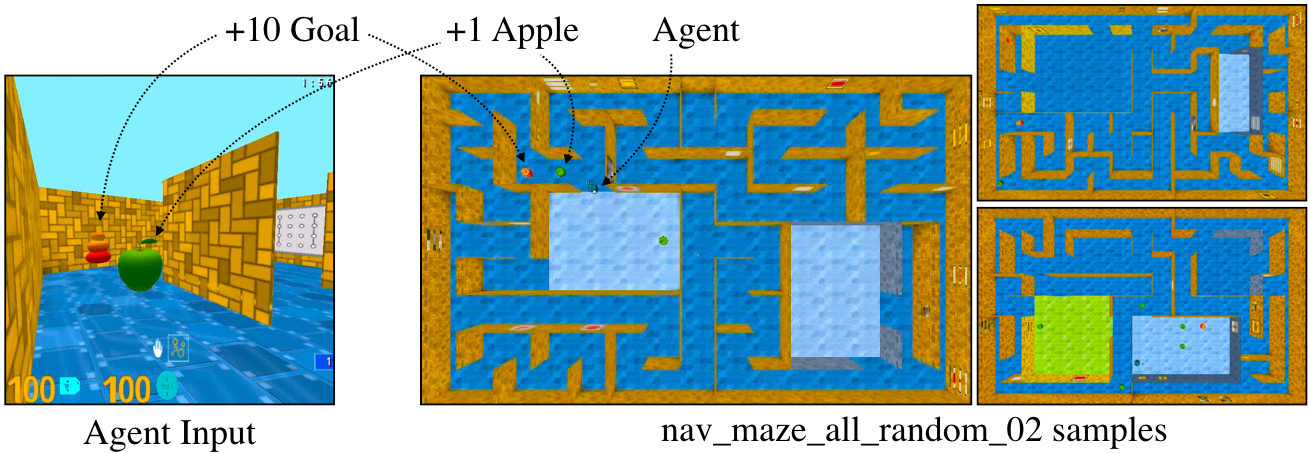}
  \caption{\small The raw RGB frame from the environment is the observation that is given as input to the agent, along with the last action and reward. This observation is shown for a sample of a maze from the $\mathrm{nav\_maze\_all\_random\_02}$ level in Labyrinth. The agent must navigate this unseen maze and pick up apples giving +1 reward and reach the goal giving +10 reward, after which it will respawn. Top down views of samples from this maze generator show the variety of mazes procedurally created. A video showing the agent playing Labyrinth levels can be viewed at \url{https://youtu.be/Uz-zGYrYEjA}}
  \label{fig:random02overview}
  \vspace{-0.5cm}
\end{figure}

In addition to learning generally about the dynamics of the environment, an agent must learn to maximise the global reward stream. To learn a policy to maximise rewards, an agent requires features that recognise states that lead to high reward and value. An agent with a good representation of rewarding states, will allow the learning of good value functions, and in turn should allow the easy learning of a policy.

However, in many interesting environments reward is encountered very sparsely, meaning that it can take a long time to train feature extractors adept at recognising states which signify the onset of reward. We want to remove the perceptual sparsity of rewards and rewarding states to aid the training of an agent, but to do so in a way which does not introduce bias to the agent's policy.

To do this, we introduce the auxiliary task of \emph{reward prediction} -- that of predicting the onset of immediate reward given some historical context. This task consists of processing a sequence of consecutive observations, and requiring the agent to predict the reward picked up in the subsequent unseen frame. This is similar to value learning focused on immediate reward ($\gamma = 0$).

Unlike learning a value function, which is used to estimate returns and as a baseline while learning a policy, the reward predictor is not used for anything other than shaping the features of the agent. This keeps us free to bias the data distribution, therefore biasing the reward predictor and feature shaping, without biasing the value function or policy.

We train the reward prediction task on sequences $S_\tau = (s_{\tau - k},s_{\tau-k+1},\ldots,s_{\tau -1})$ to predict the reward $r_\tau$, and sample $S_\tau$ from the experience of our policy $\pi$ in a skewed manner so as to over-represent rewarding events (presuming rewards are sparse within the environment). Specifically, we sample such that zero rewards and non-zero rewards are equally represented, i.e. the predicted probability of a non-zero reward is $P(r_\tau \ne 0) = 0.5$. The reward prediction is trained to minimise a loss $\mathcal{L}_\mathrm{RP}$. In our experiments we use a multiclass cross-entropy classification loss across three classes (zero, positive, or negative reward), although a mean-squared error loss is also feasible. 

The auxiliary reward predictions may use a different architecture to the agent's main policy. Rather than simply ``hanging" the auxiliary predictions off the LSTM, we use a simpler feedforward network that concatenates a stack of states $S_\tau$ after being encoded by the agent's CNN, see \figref{fig:overview} (c). The idea is to simplify the temporal aspects of the prediction task in both the future direction (focusing only on immediate reward prediction rather than long-term returns) and past direction (focusing only on immediate predecessor states rather than the complete history); the features discovered in this manner is shared with the primary LSTM (via shared weights in the convolutional encoder) to enable the policy to be learned more efficiently.

\subsection{Experience Replay}

\emph{Experience replay} has proven to be an effective mechanism for improving both the data efficiency and stability of deep reinforcement learning algorithms \citep{mnih-dqn-2015}. The main idea is to store transitions in a replay buffer, and then apply learning updates to sampled transitions from this buffer. 

Experience replay provides a natural mechanism for skewing the distribution of reward prediction samples towards rewarding events: we simply split the replay buffer into rewarding and non-rewarding subsets, and replay equally from both subsets. The skewed sampling of transitions from a replay buffer means that rare rewarding states will be oversampled, and learnt from far more frequently than if we sampled sequences directly from the behaviour policy. This approach can be viewed as a simple form of prioritised replay \citep{schaul2015prioritized}. 

In addition to reward prediction, we also use the replay buffer to perform \emph{value function replay}. This amounts to resampling recent historical sequences from the behaviour policy distribution and performing extra value function regression in addition to the on-policy value function regression in A3C. By resampling previous experience, and randomly varying the temporal position of the truncation window over which the n-step return is computed, value function replay performs value iteration and exploits newly discovered features shaped by reward prediction. We do not skew the distribution for this case. 

Experience replay is also used to increase the efficiency and stability of the auxiliary control tasks. Q-learning updates are applied to sampled experiences that are drawn from the replay buffer, allowing features to be developed extremely efficiently. 

\subsection{\emph{UNREAL} Agent}\label{sec:unreal}

The \emph{UNREAL} algorithm combines the benefits of two separate, state-of-the-art approaches to deep reinforcement learning. The primary policy is trained with A3C \citep{mnih2016asynchronous}: it learns from parallel streams of experience to gain efficiency and stability; it is updated online using policy gradient methods; and it uses a recurrent neural network to encode the complete history of experience. This allows the agent to learn effectively in partially observed environments.

The auxiliary tasks are trained on very recent sequences of experience that are stored and randomly sampled; these sequences may be prioritised (in our case according to immediate rewards) \citep{schaul2015prioritized}; these targets are trained off-policy by Q-learning; and they may use simpler feedforward architectures. This allows the representation to be trained with maximum efficiency.

The \emph{UNREAL} algorithm optimises a single combined loss function with respect to the joint parameters of the agent, $\theta$, that combines the A3C loss $\mathcal{L}_\mathrm{A3C}$ together with an auxiliary control loss $\mathcal{L}_\mathrm{PC}$, auxiliary reward prediction loss $\mathcal{L}_\mathrm{RP}$ and replayed value loss $\mathcal{L}_\mathrm{VR}$,
\begin{equation}
\mathcal{L}_\text{\emph{UNREAL}}(\theta) = \mathcal{L}_\mathrm{A3C} + \lambda_\mathrm{VR} \mathcal{L}_\mathrm{VR} + \lambda_\mathrm{PC} \sum_c \mathcal{L}_Q^{(c)} + \lambda_\mathrm{RP} \mathcal{L}_\mathrm{RP}
\vspace{-0.25cm}
\end{equation}

where $\lambda_\mathrm{VR}, \lambda_\mathrm{PC}, \lambda_\mathrm{RP}$ are weighting terms on the individual loss components.

In practice, the loss is broken down into separate components that are applied either on-policy, directly from experience; or off-policy, on replayed transitions. Specifically, the A3C loss $\mathcal{L}_\mathrm{A3C}$ is minimised on-policy; while the value function loss $\mathcal{L}_\mathrm{VR}$ is optimised from replayed data, in addition to the A3C loss (of which it is one component, see \sref{sec:background}). The auxiliary control loss $\mathcal{L}_\mathrm{PC}$ is optimised off-policy from replayed data, by $n$-step Q-learning. Finally, the reward loss $\mathcal{L}_\mathrm{RP}$ is optimised from rebalanced replay data.

\section{Experiments}\label{sec:exp}

\begin{figure}[t]
\centering
\vspace{-1.2cm}
\hspace{-0.5cm}
\centering
\begin{tabular}{cc}
{\small Labyrinth Performance} & {\small Labyrinth Robustness}\\
\includegraphics[scale=0.3]{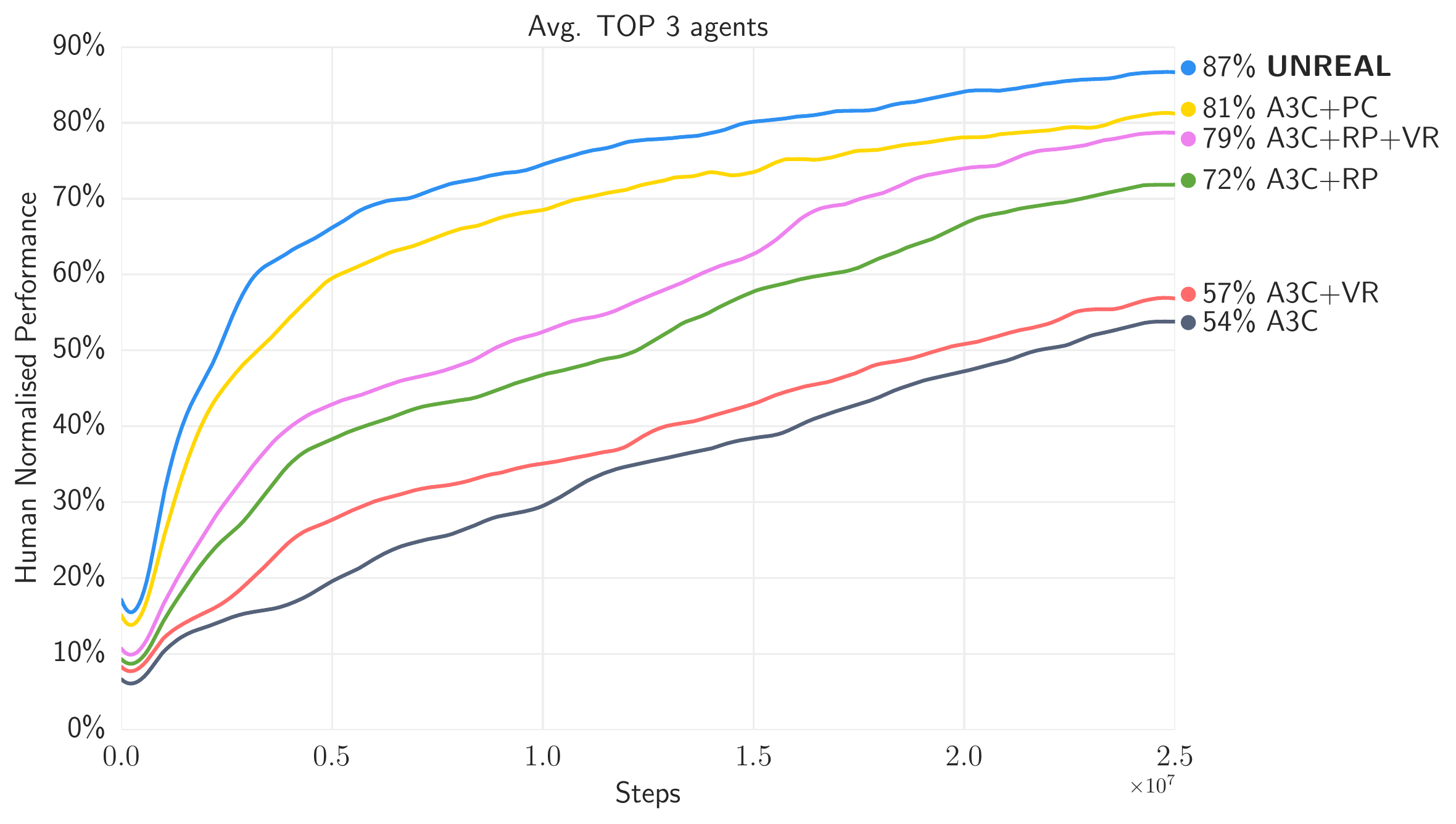}&
\includegraphics[scale=0.3]{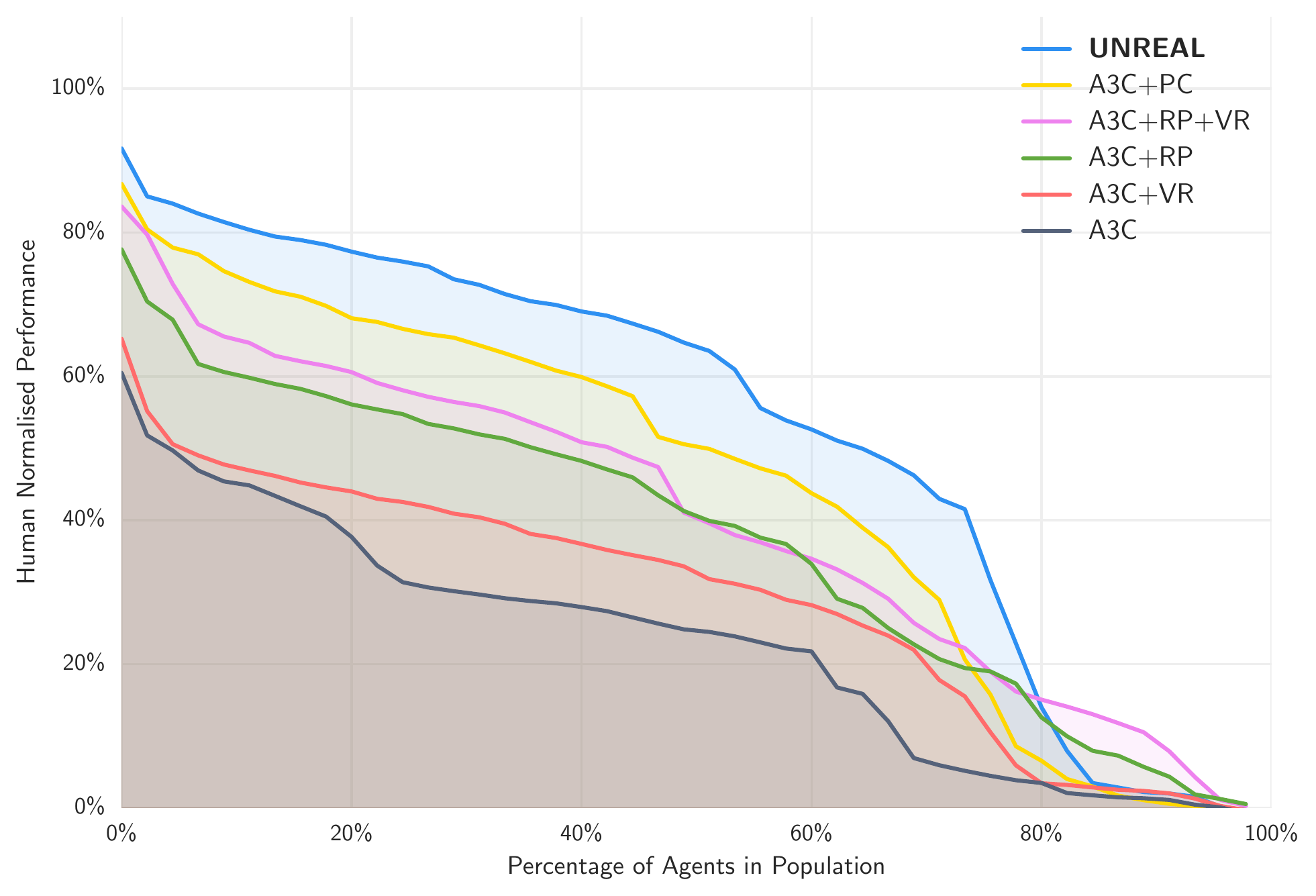}\\
\\
{\small Atari Performance} & {\small Atari Robustness}\\
\includegraphics[scale=0.3]{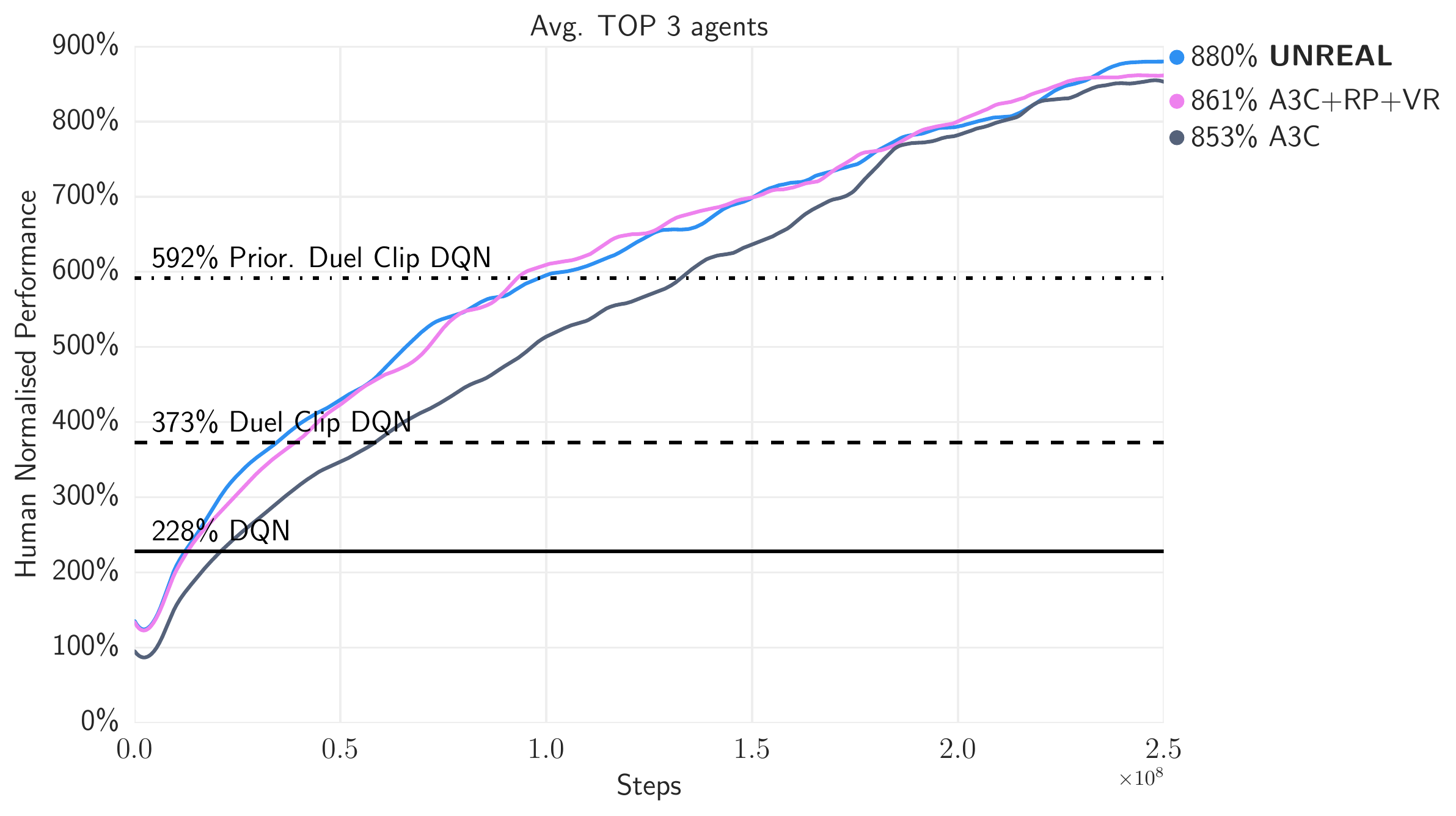}&
\includegraphics[scale=0.3]{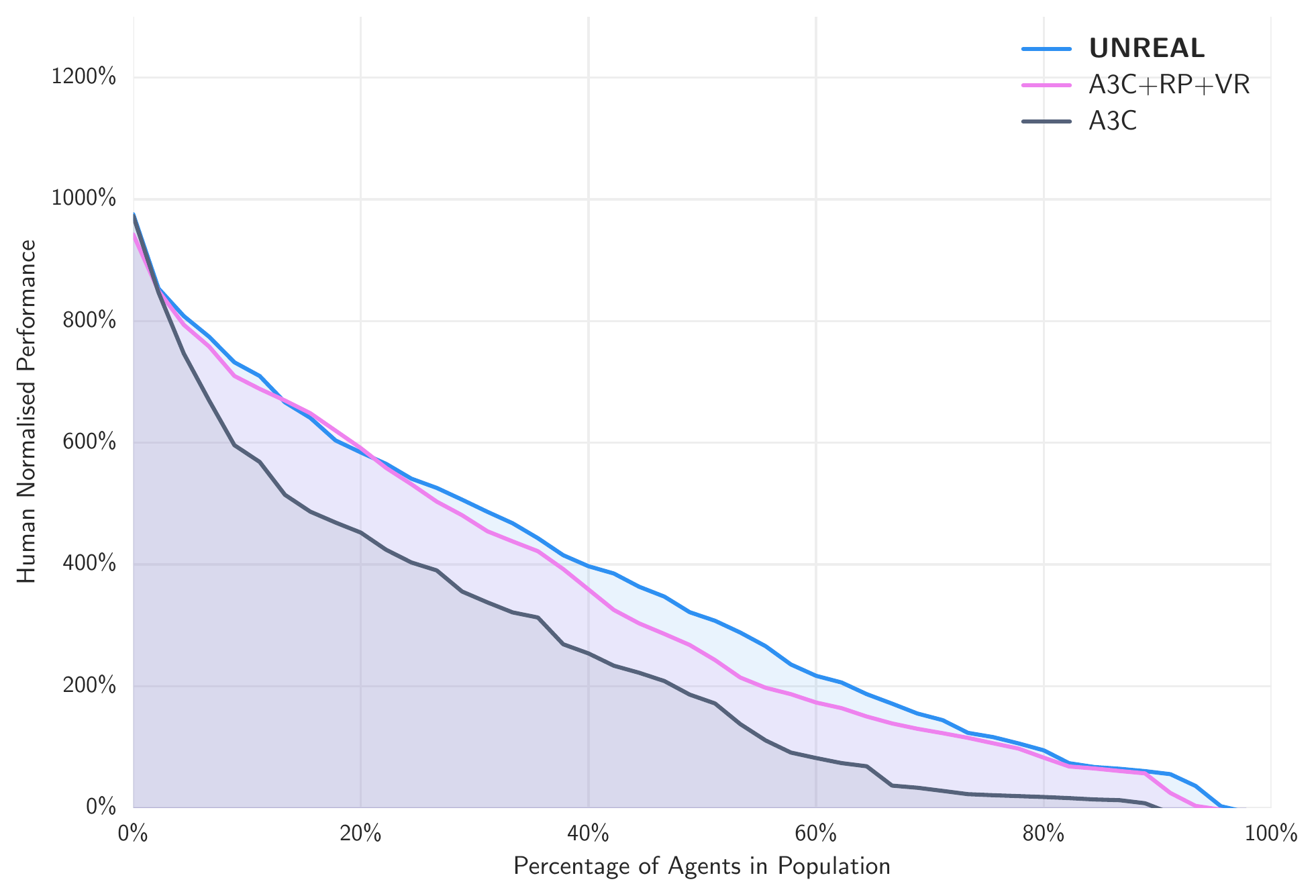}
\end{tabular}
  \caption{\small An overview of performance averaged across all levels on Labyrinth (Top) and Atari (Bottom). In the ablated versions RP is reward prediction, VR is value function replay, and PC is pixel control, with the \emph{UNREAL} agent being the combination of all. \emph{Left}: The mean human-normalised performance over last 100 episodes of the top-3 jobs at every point in training. We achieve an average of 87\% human-normalised score, with every element of the agent improving upon the 54\% human-normalised score of vanilla A3C. \emph{Right}: The final human-normalised score of every job in our hyperparameter sweep, sorted by score. On both Labyrinth and Atari, the \emph{UNREAL} agent increases the robustness to the hyperparameters (namely learning rate and entropy cost).}
\label{fig:performance}
\vspace{-0.5cm}
\end{figure}

In this section we give the results of experiments performed on the 3D environment \emph{Labyrinth} in \sref{sec:lab} and Atari in \sref{sec:atari}.

In all our experiments we used an A3C CNN-LSTM agent as our baseline and the \emph{UNREAL} agent along with its ablated variants added auxiliary outputs and losses to this base agent. The agent is trained on-policy with 20-step returns and the auxiliary tasks are performed every 20 environment steps, corresponding to every update of the base A3C agent. The replay buffer stores the most recent 2k observations, actions, and rewards taken by the base agent. In Labyrinth we use the same set of 17 discrete actions for all games and on Atari the action set is game dependent (between 3 and 18 discrete actions). The full implementation details can be found in \sref{sec:details}.

\subsection{Labyrinth Results}\label{sec:lab}

Labyrinth is a first-person 3D game platform extended from OpenArena \citep{OpenArena}, which is itself based on Quake3 \citep{QuakeThree}. Labyrinth is comparable to other first-person 3D game platforms for AI research like VizDoom \citep{kempka2016vizdoom} or Minecraft~\citep{tessler2016deep}. However, in comparison, Labyrinth has considerably richer visuals and more realistic physics. Textures in Labyrinth are often dynamic (animated) so as to convey a game world where walls and floors shimmer and pulse, adding significant complexity to the perceptual task. The action space allows for fine-grained pointing in a fully 3D world. Unlike in VizDoom, agents can look up to the sky or down to the ground. Labyrinth also supports continuous motion unlike the Minecraft platform of \citep{oh2016control}, which is a 3D grid world.

We evaluated agent performance on 13 Labyrinth levels that tested a range of different agent abilities. 
A top-down visualization showing the layout of each level can be found in \figref{fig:levels} of the Appendix. A gallery of example images from the first-person perspective of the agent are in \figref{fig:first_person_views} of the Appendix. The levels can be divided into four categories:

\begin{enumerate}
\item Simple fruit gathering levels with a static map ($\mathrm{seekavoid\_arena\_01}$ and $\mathrm{stairway\_to\_melon\_01}$). The goal of these levels is to collect apples (small positive reward) and melons (large positive reward)  while avoiding lemons (small negative reward).

\item Navigation levels with a static map layout ($\mathrm{nav\_maze\_static\_0\{1,2,3\}}$ and $\mathrm{nav\_maze\_random\_goal\_0\{1,2,3\}}$). These levels test the agent's ability to find their way to a goal in a fixed maze that remains the same across episodes. The starting location is random. In this case, agents could encode the structure of the maze in network weights. In the random goal variant, the location of the goal changes in every episode. The optimal policy is to find the goal's location at the start of each episode and then use long-term knowledge of the maze layout to return to it as quickly as possible from any location. The static variant is simpler in that the goal location is always fixed for all episodes and only the agent's starting location changes so the optimal policy does not require the first step of exploring to find the current goal location.

\item Procedurally-generated navigation levels requiring effective exploration of a new maze generated on-the-fly at the start of each episode ($\mathrm{nav\_maze\_all\_random\_0\{1,2,3\}}$). These levels test the agent's ability to effectively explore a totally new environment. The optimal policy would begin by exploring the maze to rapidly learn its layout and then exploit that knowledge to repeatedly return to the goal as many times as possible before the end of the episode (between 60 and 300 seconds).

\item Laser-tag levels requiring agents to wield laser-like science fiction gadgets to tag bots controlled by the game's in-built AI ($\mathrm{lt\_horse\_shoe\_color}$ and $\mathrm{lt\_hallway\_slope}$). A reward of $1$ is delivered whenever the agent tags a bot by reducing its shield to 0. These levels approximate the default  OpenArena/Quake3 gameplay mode. In $\mathrm{lt\_hallway\_slope}$ there is a sloped arena, requiring the agent to look up and down. In  $\mathrm{lt\_horse\_shoe\_color}$, the colors and textures of the bots are randomly generated at the start of each episode. This prevents agents from relying on color for bot detection. These levels test aspects of fine-control (for aiming), planning (to anticipate where bots are likely to move), strategy (to control key areas of the map such as gadget spawn points), and robustness to the substantial visual complexity arising from the large numbers of independently moving objects (gadget projectiles and bots). 
\end{enumerate}

\subsubsection{Results}

We compared the full \emph{UNREAL} agent to a basic A3C LSTM agent along with several ablated versions of \emph{UNREAL} with different components turned off. A video of the final agent performance, as well as visualisations of the activations and auxiliary task outputs can be viewed at \url{https://youtu.be/Uz-zGYrYEjA}.

\figref{fig:performance} (right) shows curves of mean human-normalised scores over the 13 Labyrinth levels. Adding each of our proposed auxiliary tasks to an A3C agent substantially improves the performance. Combining different auxiliary tasks leads to further improvements over the individual auxiliary tasks. The \emph{UNREAL} agent, which combines all three auxiliary tasks, achieves more than twice the final human-normalised mean performance of A3C, increasing from 54\% to 87\% (45\% to 92\% for median performance). This includes a human-normalised score of 116\% on $\mathrm{lt\_hallway\_slope}$ and 100\% on $\mathrm{nav\_maze\_random\_goal\_02}$.

\begin{figure}[t]
\centering
\vspace{-1.2cm}
\hspace{-5mm} \includegraphics[scale=0.35]{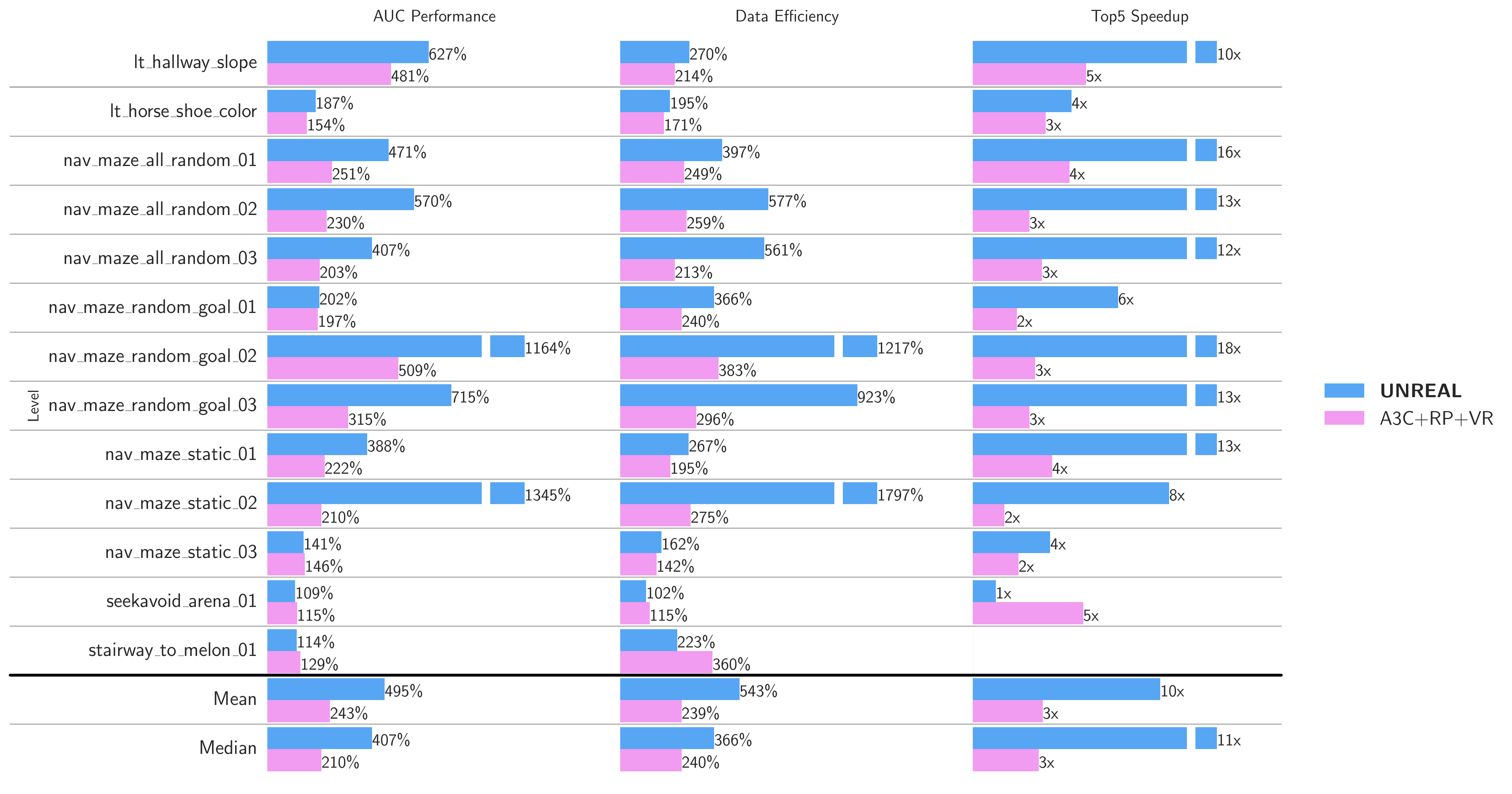}
  \caption{\small A breakdown of the improvement over A3C due to our auxiliary tasks for each level on Labyrinth. The values for A3C+RP+VR (reward prediction and value function replay) and \emph{UNREAL} (reward prediction, value function replay and pixel control) are normalised by the A3C value. AUC Performance gives the robustness to hyperparameters (area under the robustness curve \figref{fig:performance} Right). Data Efficiency is area under the mean learning curve for the top-5 jobs, and Top5 Speedup is the speedup for the mean of the top-5 jobs to reach the maximum top-5 mean score set by A3C. Speedup is not defined for $\mathrm{stairway\_to\_melon}$ as A3C did not learn throughout training.}
  \label{fig:efficiency}
  \vspace{-0.3cm}
\end{figure}

Perhaps of equal importance, aside from final performance on the games, \emph{UNREAL} is significantly faster at learning and therefore more data efficient, achieving a mean speedup of the number of steps to reach A3C best performance of 10$\times$ (median 11$\times$) across all levels and up to 18$\times$ on $\mathrm{nav\_maze\_random\_goal\_02}$. This translates in a drastic improvement in the data efficiency of \emph{UNREAL} over A3C, requiring less than 10\% of the data to reach the final performance of A3C. We can also measure the robustness of our learning algorithms to hyperparameters by measuring the performance over all hyperparameters (namely learning rate and entropy cost). This is shown in \figref{fig:performance} Top: every auxiliary task in our agent improves robustness. A breakdown of the performance of A3C, \emph{UNREAL} and \emph{UNREAL} without pixel control on the individual Labyrinth levels is shown in \figref{fig:efficiency}. 

\begin{figure}[t]
\centering
\includegraphics[scale=0.25]{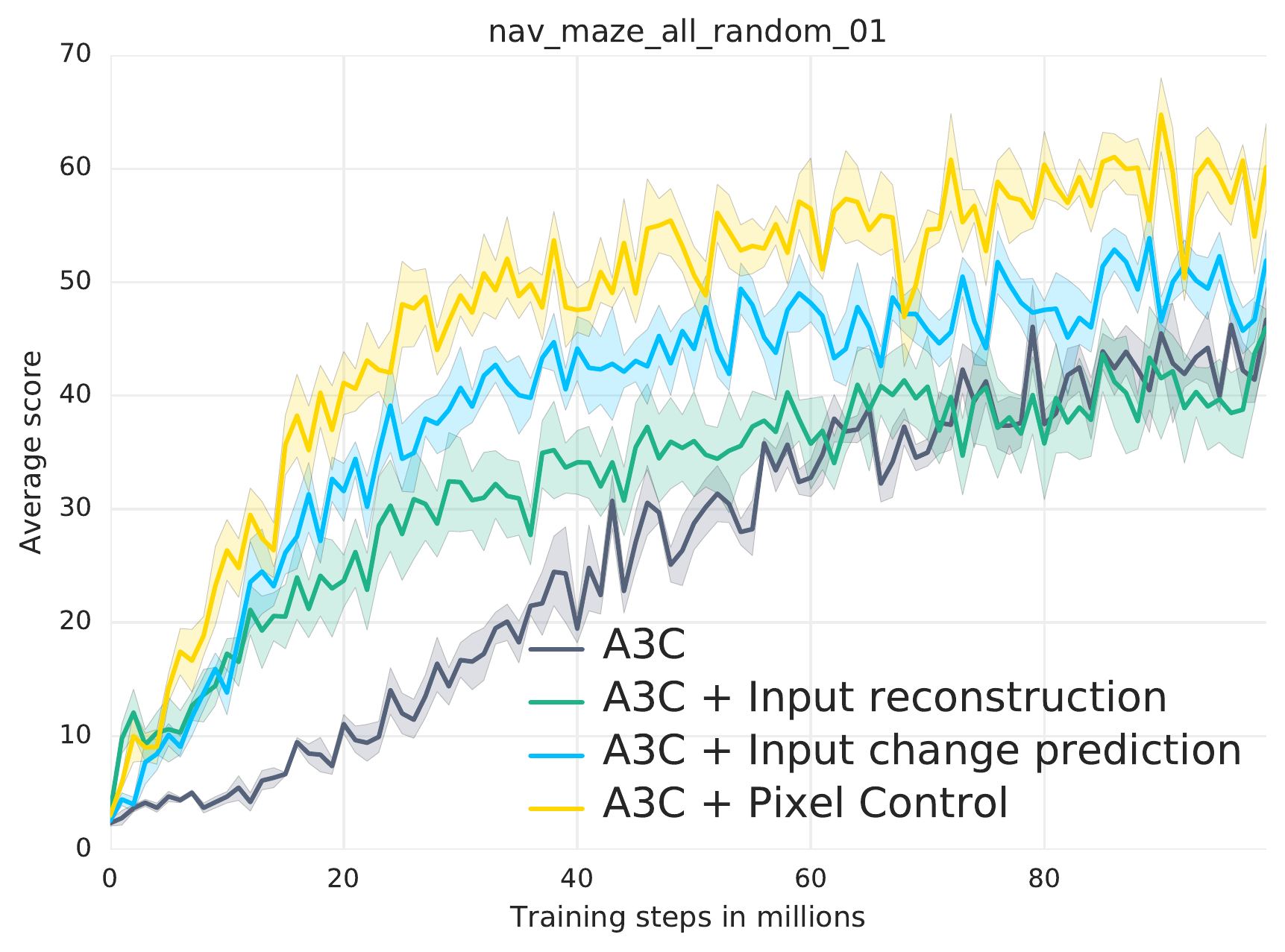}
\includegraphics[scale=0.25]{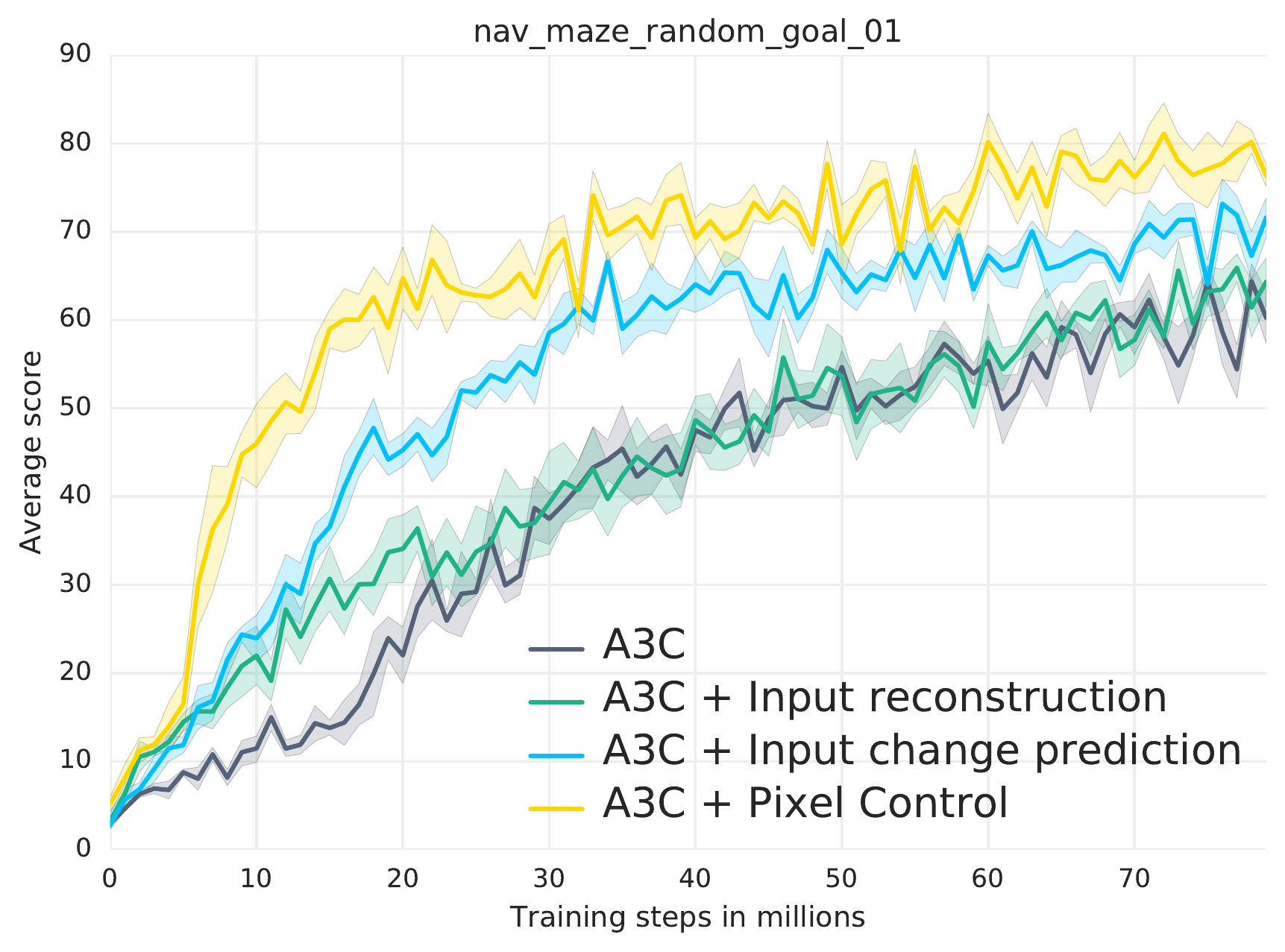}
\includegraphics[scale=0.25]{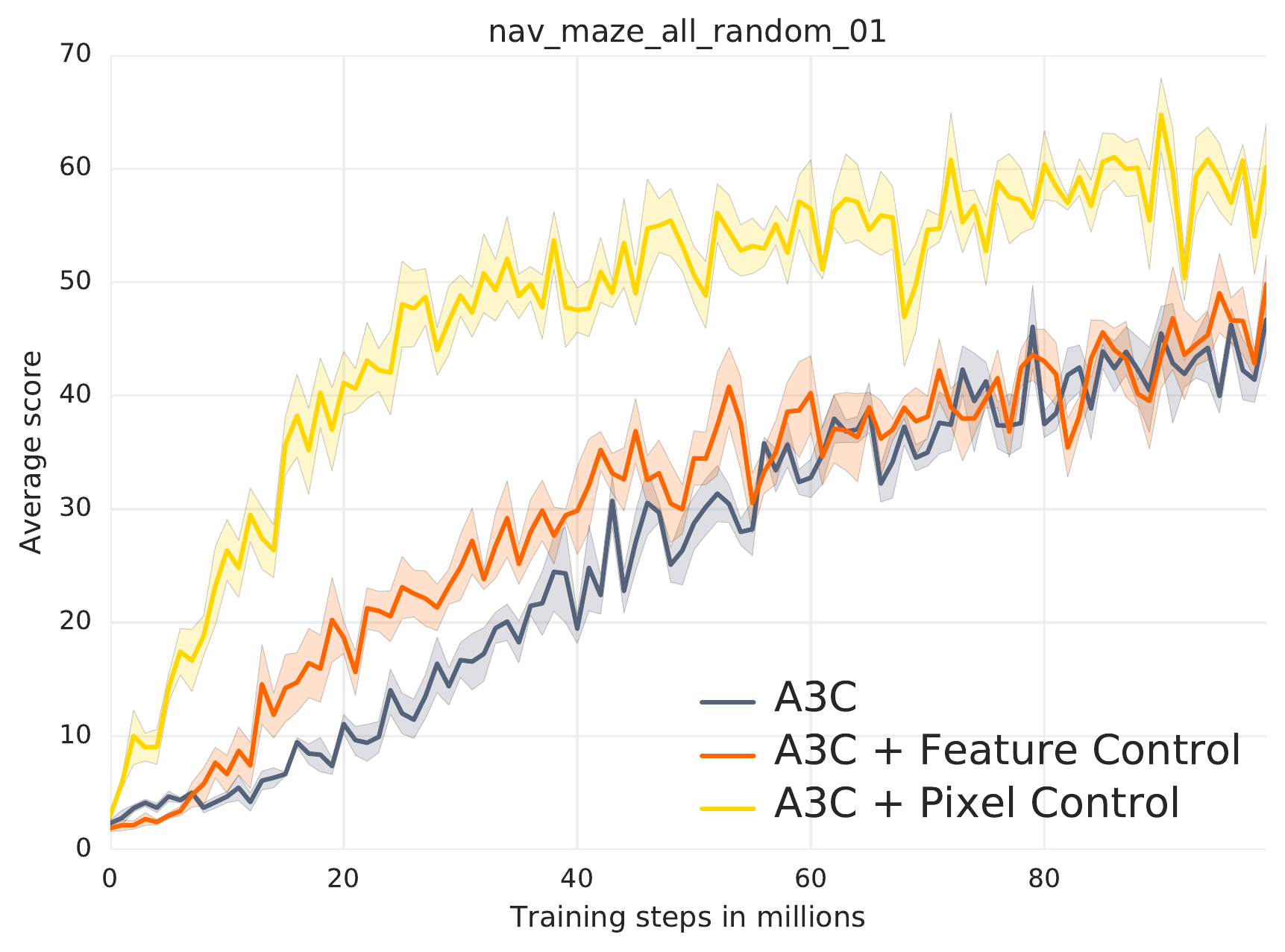}
  \caption{\small Comparison of various forms of self-supervised learning on random maze navigation. Adding an input reconstruction loss to the objective leads to faster learning compared to an A3C baseline. Predicting changes in the inputs works better than simple image reconstruction. Learning to control changes leads to the best results. \label{fig:pxc-baselines} }
  \vspace{-0.4cm}
\end{figure}

\paragraph{Unsupervised Reinforcement Learning}
In order to better understand the benefits of auxiliary control tasks we compared it to two simple baselines on three Labyrinth levels. The first baseline was A3C augmented with a pixel reconstruction loss, which has been shown to improve performance on 3D environments \citep{kulkarni2016deep}. The second baseline was A3C augmented with an input change prediction loss, which can be seen as simply predicting the immediate auxiliary reward instead of learning to control. Finally, we include preliminary results for A3C augmented with the feature control auxiliary task on one of the levels. We retuned the hyperparameters of all methods (including learning rate and the weight placed on the auxiliary loss) for each of the three Labyrinth levels. \figref{fig:pxc-baselines} shows the learning curves for the top 5 hyperparameter settings on three Labyrinth navigation levels. The results show that learning to control pixel changes is indeed better than simply predicting immediate pixel changes, which in turn is better than simply learning to reconstruct the input. In fact, learning to reconstruct only led to faster initial learning and actually made the final scores worse when compared to vanilla A3C. Our hypothesis is that input reconstruction hurts final performance because it puts too much focus on reconstructing irrelevant parts of the visual input instead of visual cues for rewards, which rewarding objects are rarely visible. Encouragingly, we saw an improvement from including the feature control auxiliary task. Combining feature control with other auxiliary tasks is a  promising future direction.

\subsection{Atari}\label{sec:atari}

We applied the \emph{UNREAL} agent as well as \emph{UNREAL} without pixel control to 57 Atari games from the Arcade Learning Environment ~\citep{bellemare-ale} domain. We use the same evaluation protocol as for our Labyrinth experiments where we evaluate 50 different random hyper parameter settings (learning rate and entropy cost) on each game. The results are shown in the bottom row of Figure \ref{fig:performance}. The left side shows the average performance curves of the top 3 agents for all three methods the right half shows sorted average human-normalised scores for each hyperparameter setting. More detailed learning curves for individual levels can be found in \figref{fig:levels}. We see that \emph{UNREAL} surpasses the current state-of-the-art agents, \ie~A3C and Prioritized Dueling DQN~\citep{wang2015dueling}, across all levels attaining 880\% mean and 250\% median performance. Notably, \emph{UNREAL} is also substantially more robust to hyper parameter settings than A3C.

\section{Conclusion}
We have shown how augmenting a deep reinforcement learning agent with auxiliary control and reward prediction tasks can drastically improve both data efficiency and robustness to hyperparameter settings. Most notably, our proposed \emph{UNREAL} architecture more than doubled the previous state-of-the-art results on the challenging set of 3D Labyrinth levels, bringing the average scores to over $87\%$ of human scores. The same \emph{UNREAL} architecture also significantly improved both the learning speed and the robustness of A3C over 57 Atari games.

\section*{Acknowledgements}
We thank Charles Beattie, Julian Schrittwieser, Marcus Wainwright, and Stig Petersen for environment design and development,
and Amir Sadik and Sarah York for expert human game testing. We also thank Joseph Modayil, Andrea Banino, Hubert Soyer, Razvan Pascanu, and Raia Hadsell for many helpful discussions.

\bibliographystyle{iclr2017_conference}
\bibliography{references}

\begin{appendices}

\clearpage
\section{Atari Games}
\vspace{-0.6cm}
\begin{figure}[h!]
\centering
\includegraphics[width=0.32\textwidth]{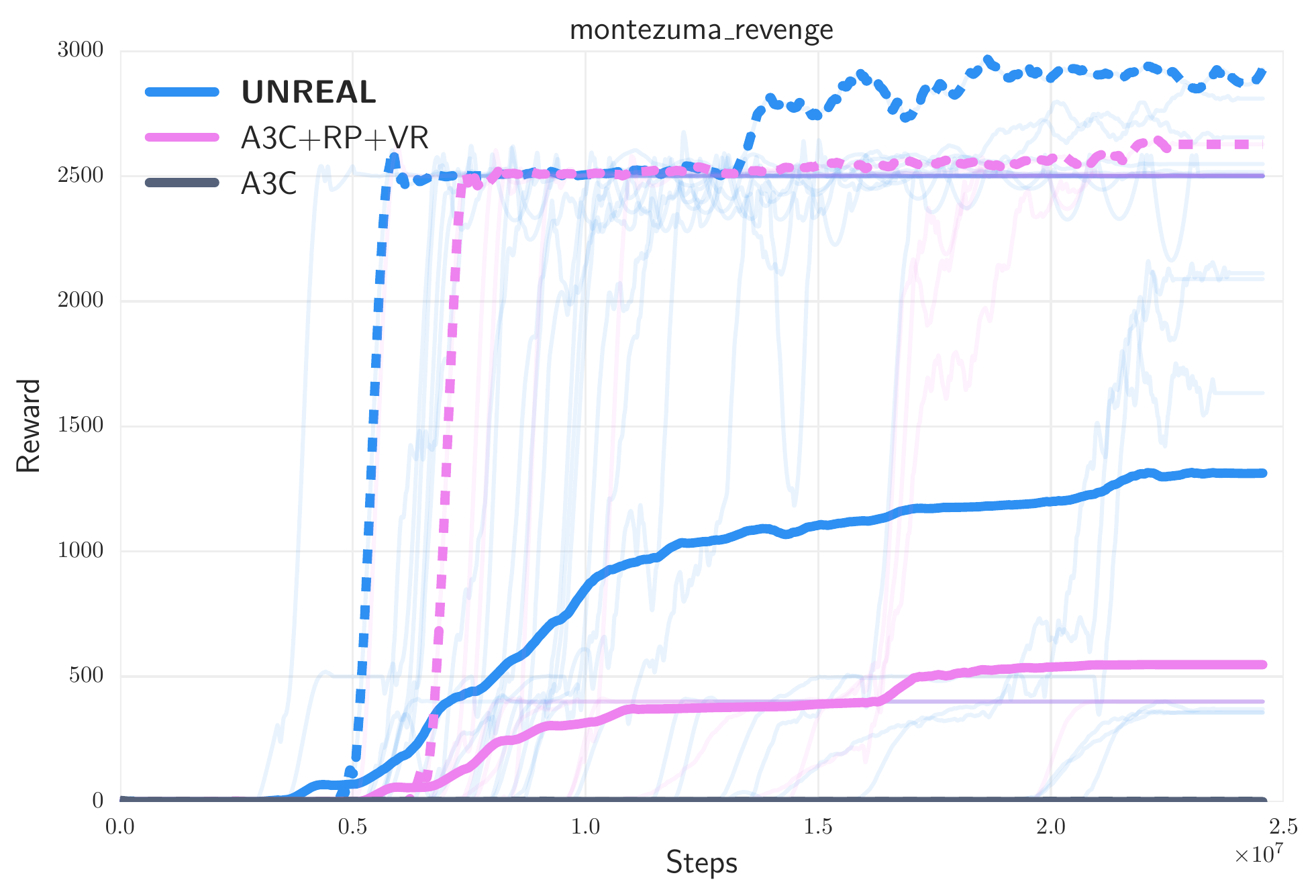}
\includegraphics[width=0.32\textwidth]{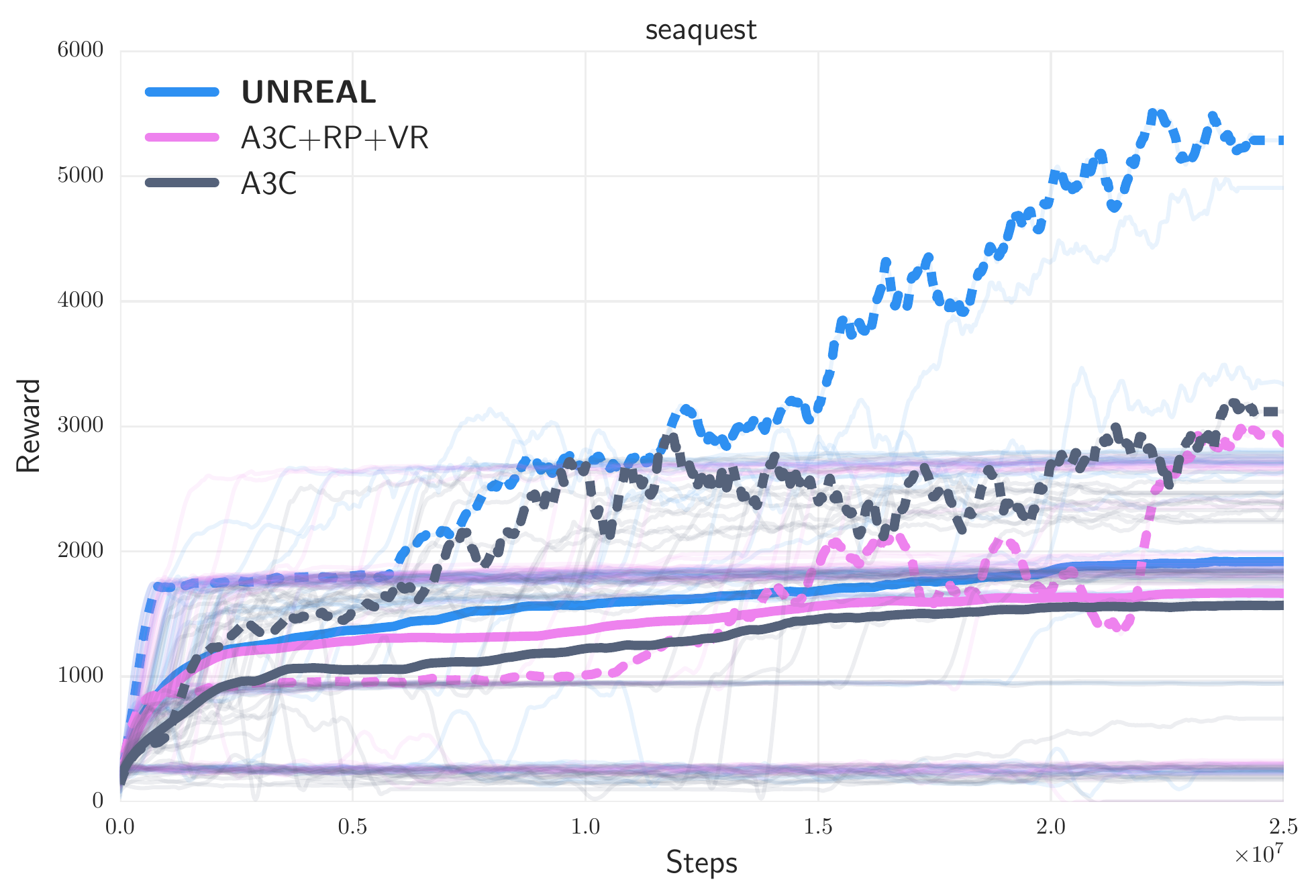} 
\includegraphics[width=0.32\textwidth]{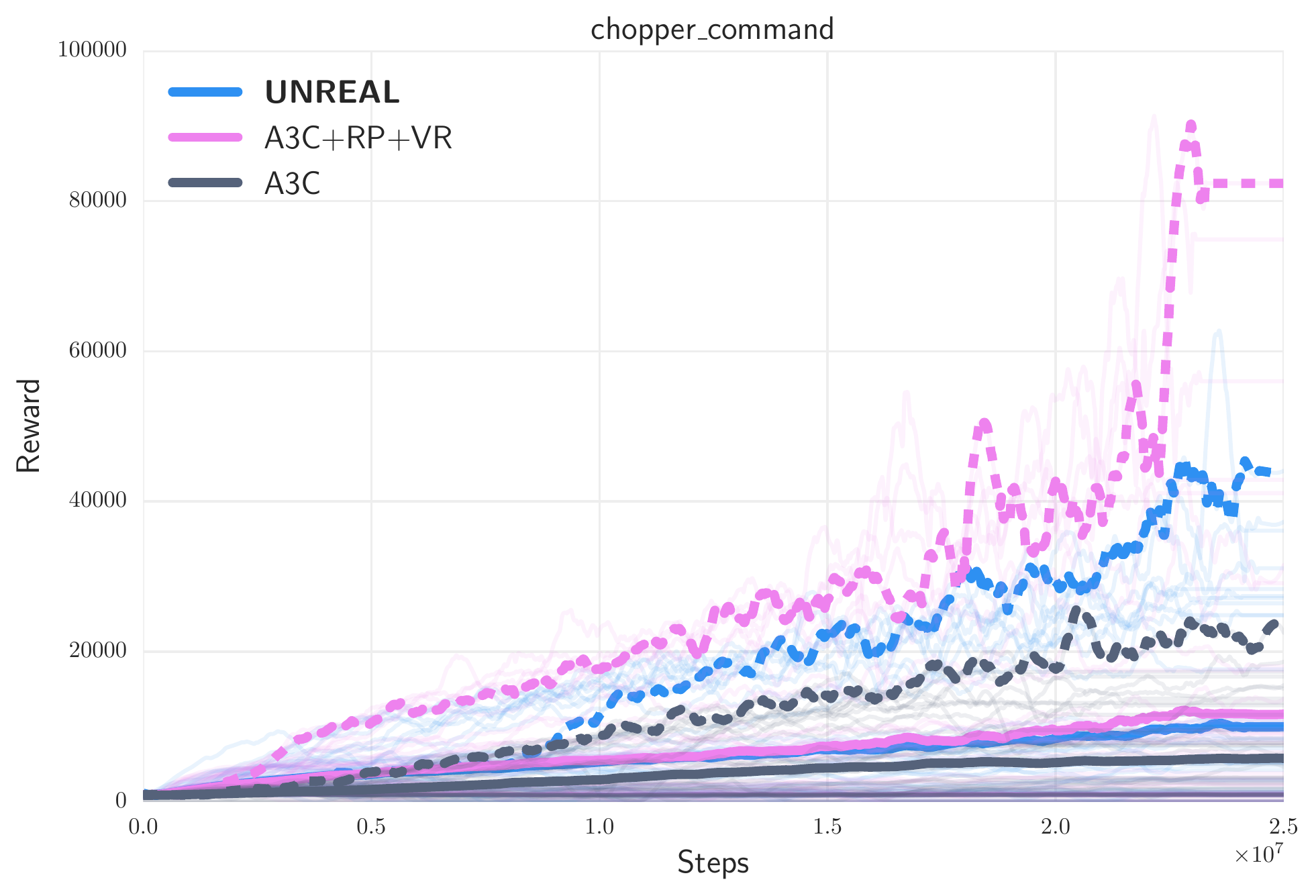}
\caption{Learning curves for three example Atari games. Semi-transparent lines are agents with different seeds and hyperparameters, the bold line is a mean over population and dotted line is the best agent (in terms of final performance).}
\label{fig:levels}
\vspace{-0.7cm}
\end{figure}

\section{Implementation Details}\label{sec:details}
The input to the agent at each timestep was an $84\times84$ RGB image. All agents processed the input with the convolutional neural network (CNN) originally used for Atari by \cite{mnih-atari-2013}. The network consists of two convolutional layers. The first one has $16$ $8\times 8$ filters applied with stride $4$, while the second one has $32$ $4\times4$ filters with stride $2$. This is followed by a fully connected layer with $256$ units. All three layers are followed by a ReLU non-linearity. All agents used an LSTM  with forget gates \citep{gers2000learning} with 256 cells which take in the CNN-encoded observation concatenated with the previous action taken and curren:t reward. The policy and value function are linear projections of the LSTM output. The agent is trained with 20-step unrolls. The action space of the agent in the environment is game dependent for Atari (between 3 and 18 discrete actions), and 17 discrete actions for Labyrinth. Labyrinth runs at 60 frames-per-second. We use an action repeat of four, meaning that each action is repeated four times, with the agent receiving the final fourth frame as input to the next processing step. 

For the pixel control auxiliary tasks we trained policies to control the central $80\times80$ crop of the inputs. The cropped region was subdivided into a $20\times 20$ grid of non-overlapping $4\times 4$ cells. The instantaneous reward in each cell was defined as the average absolute difference from the previous frame, where the average is taken over both pixels and channels in the cell. The output tensor of auxiliary values, $Q^\mathrm{aux}$, is produced from the LSTM outputs by a deconvolutional network. The LSTM outputs are first mapped to a $32\times7\times7$ spatial feature map with a linear layer followed by a ReLU.  Deconvolution layers with $1$ and $N_\mathrm{act}$ filters of size $4\times 4$ and stride 2 map the $32\times7\times7$ into a value tensor and an advantage tensor respectively. The spatial map is then decoded into Q-values using the dueling parametrization \citep{wang2015dueling} producing the $N_\mathrm{act}\times20\times20$ output $Q^\mathrm{aux}$.

The architecture for feature control was similar. We learned to control the second hidden layer, which is a spatial feature map with size $32\times9 \times 9$. Similarly to pixel control, we exploit the spatial structure in the data and used a deconvolutional network to produce $Q^\mathrm{aux}$ from the LSTM outputs. Further details are included in the supplementary materials.

The reward prediction task is performed on a sequence of three observations, which are fed through three instances of the agent's CNN. The three encoded CNN outputs are concatenated and fed through a fully connected layer of 128 units with ReLU activations, followed by a final linear three-class classifier and softmax. The reward is predicted as one of three classes: positive, negative, or zero and trained with a task weight $\lambda_\mathrm{RP}=1$. The value function replay is performed on a sequence of length 20 with a task weight $\lambda_\mathrm{VR}=1$.

The auxiliary tasks are performed every 20 environment steps, corresponding to every update of the base A3C agent, once the replay buffer has filled with agent experience. The replay buffer stores the most recent 2k observations, actions, and rewards taken by the base agent.

The agents are optimised over 32 asynchronous threads with shared RMSprop \citep{mnih2016asynchronous}. The learning rates are sampled from a log-uniform distribution between 0.0001 and 0.005. The entropy costs are sampled from the log-uniform distribution between 0.0005 and 0.01. Task weight $\lambda_\mathrm{PC}$ is sampled from log-uniform distribution between 0.01 and 0.1 for Labyrinth and 0.0001 and 0.01 for Atari (since Atari games are not homogeneous in terms of pixel intensities changes, thus we need to fit this normalization factor).

\section{Labyrinth Levels}
\begin{figure}[h]
\centering
\begin{tabular}{cc}
\includegraphics[width=0.35\textwidth]{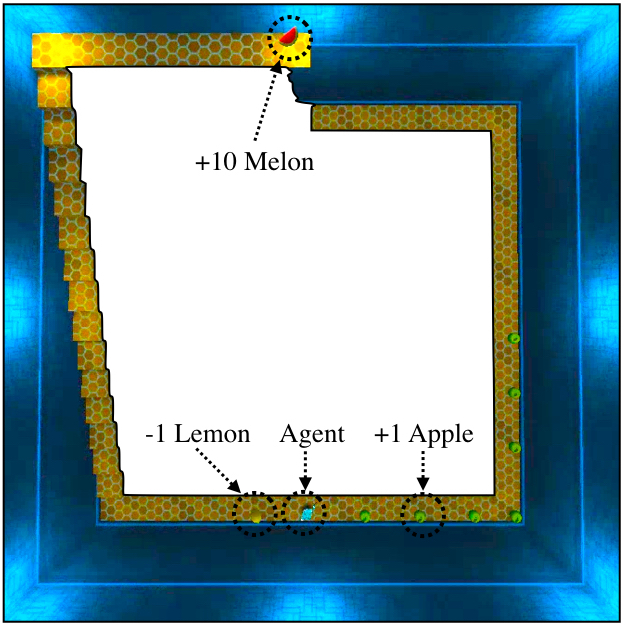} & \includegraphics[width=0.35\textwidth]{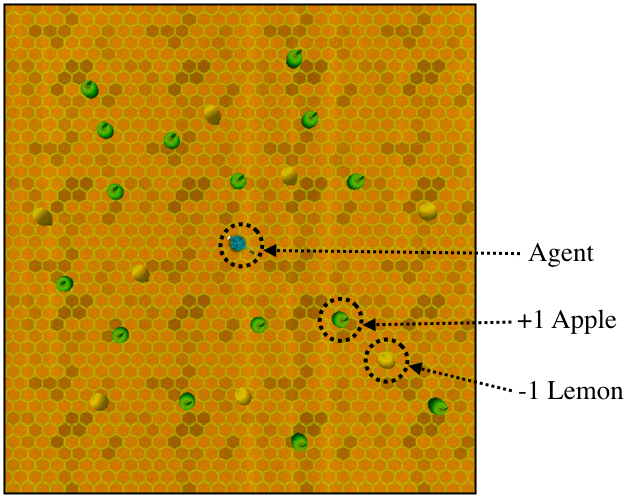}\\
$\mathrm{stairway\_to\_melon}$ & $\mathrm{seekavoid\_arena\_01}$ \\
\hline \\
\includegraphics[width=0.35\textwidth]{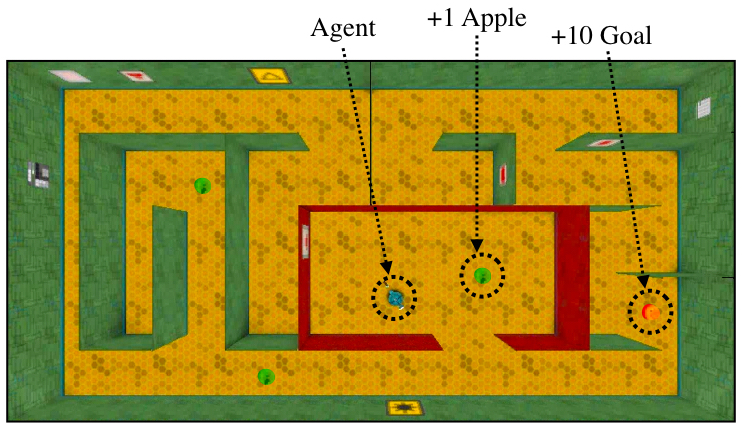} & \includegraphics[width=0.35\textwidth]{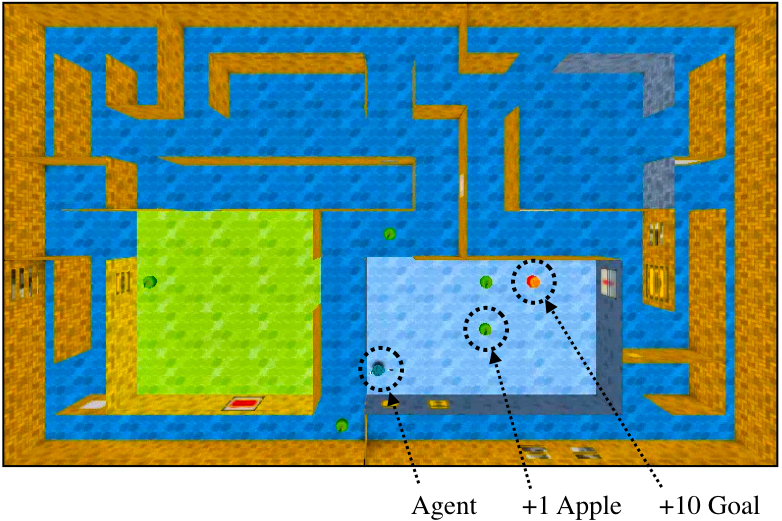}\\
$\mathrm{nav\_maze*01}$ & $\mathrm{nav\_maze*02}$ \\
\hline \\
\includegraphics[width=0.45\textwidth]{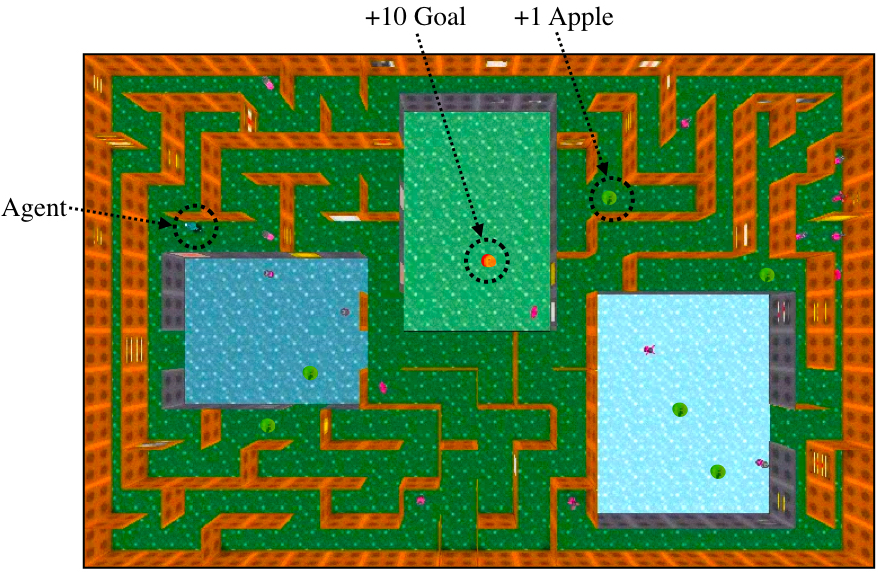} & \includegraphics[width=0.3\textwidth]{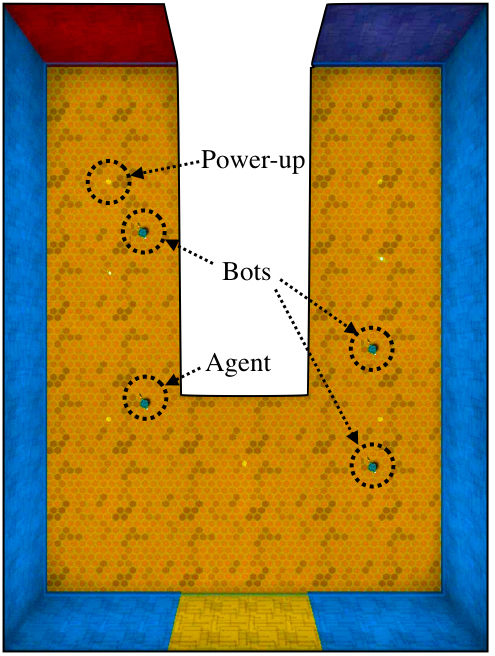}\\
$\mathrm{nav\_maze*03}$ & $\mathrm{lt\_horse\_shoe\_color}$ \\
\hline \\
\multicolumn{2}{c}{\includegraphics[width=0.5\textwidth]{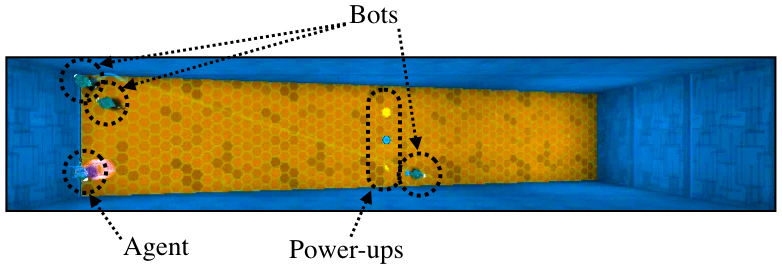}}\\
\multicolumn{2}{c}{$\mathrm{lt\_hallway\_slope}$}\\
\end{tabular}
\caption{Top-down renderings of each Labyrinth level. The $\mathrm{nav\_maze*\_0\{1,2,3\}}$ levels show one example maze layout. In the $\mathrm{all\_random}$ case, a new maze was randomly generated at the start of each episode.}
\label{fig:levels}
\end{figure}

\begin{figure}[t]
\centering
\includegraphics[scale = .8]{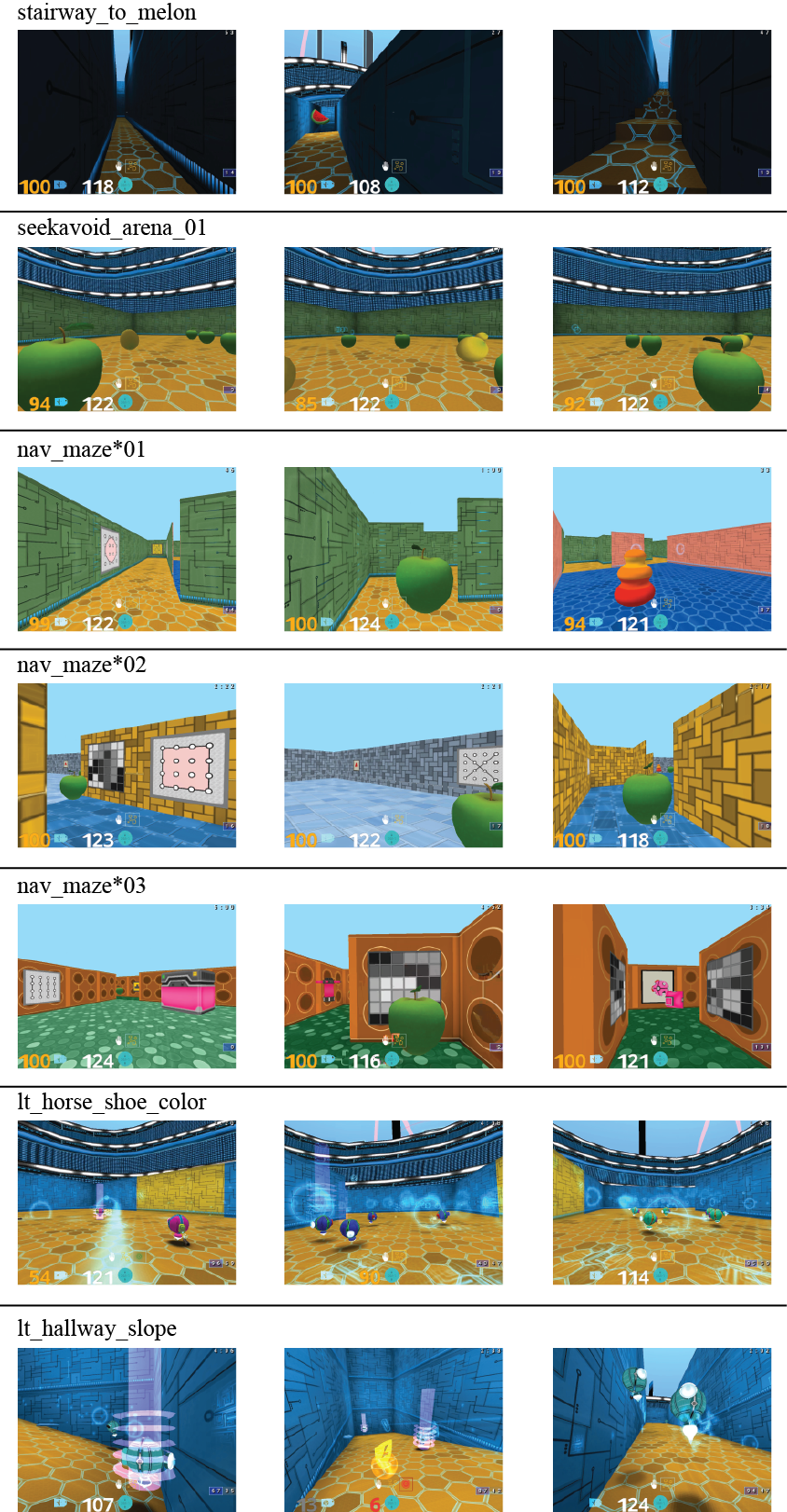}
  \caption{\small Example images from the agent's egocentric viewpoint for each Labyrinth level. \label{fig:first_person_views} }
\end{figure}

\end{appendices}

\end{document}